\documentclass[10pt]{article}
\usepackage{rbpreprint}
\usepackage[authoryear,round]{natbib}
\setcitestyle{authoryear,round,citesep={;},aysep={,},yysep={;}}
\usepackage{graphicx}
\usepackage{textcomp}
\usepackage{booktabs}
\usepackage{tabularx}
\usepackage{enumitem}
\usepackage[colorlinks=true,linkcolor=rblink,citecolor=rblink,urlcolor=rblink]{hyperref}
\usepackage{url}
\usepackage{placeins}

\hypersetup{
  pdftitle={Razavi-Bench: Tracking AI Progress in Analog Circuit Reasoning},
  pdfauthor={Zhishuai Zhang, Oscar Toomey, Taekwang Jang},
}

\title{Razavi-Bench: Tracking AI Progress in\\
Analog Circuit Reasoning}
\byline{Zhishuai Zhang, Oscar Toomey, Taekwang Jang\\
{\mdseries ETH Zurich}}
\abstracttext{%
We propose Razavi-Bench, a benchmark built from the 50 questions in Razavi's two
\emph{Analog Design Experiments With AI} articles and graded by GPT-6 Luna against
reference answers and rubrics, and use it to track how fast AI models improve at
analog circuit analysis.  Across 73 multimodal large language models released
between May 2024 and November 2026, the top score rose from 28.83\% to 98.33\% in 30
months, mostly before the reference answers were published.  The leading models now
answer almost every textbook question correctly; the hardest questions show a
circuit that resembles a textbook one but works differently, or ask about behavior
the obvious analysis misses.  Within two and a half years, a set of questions that
exposed the limits of AI models has become one they answer almost perfectly.}
\metadata{Project Page}{\url{https://razavi-bench.tokenzhang.com}}
\metadata{Repository}{\url{https://github.com/Arcadia-1/razavi-bench}}

\begin{document}
\maketitle

\section{Can AI Models Analyze Analog Circuits?}
\label{sec:intro}

How well can an AI model reason about analog circuits?  Razavi's two \emph{Analog
Design Experiments With AI} articles asked whether a model can interpret a
schematic, identify the mechanisms at work, and predict the effect of a design
change.  ChatGPT answered fewer than half of his undergraduate-level analysis
questions correctly \citep{razavi2025analogai1}, and Gemini 3.1 Pro just over half
of his design-oriented ones \citep{razavi2026analogai2}.  Both misread devices or
signal paths, or applied valid formulas under the wrong assumptions.

\begin{figure}[t]
\centering
\includegraphics[width=\textwidth]{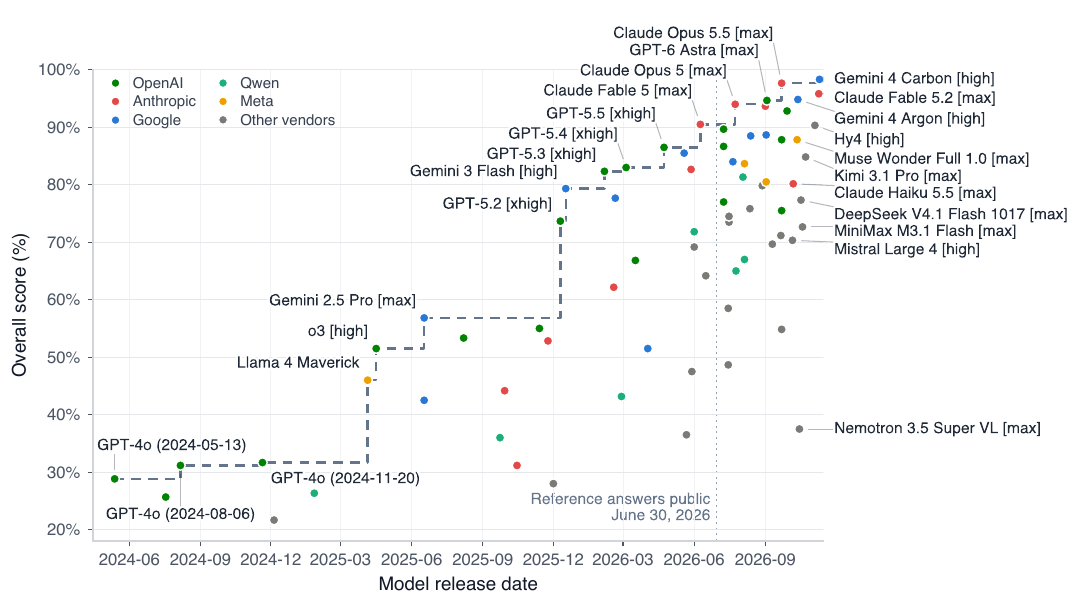}
\caption{Overall scores of all 73 models at their highest reasoning effort, by
release date.  The dashed step line is the top score available at each date; the
dotted line marks June 30, 2026, when the full reference answers became public.}
\label{fig:timeline}
\end{figure}

These are two snapshots of a fast-moving field.  Chip-design benchmarks such as
VerilogEval \citep{liu2023verilogeval} and RTLLM \citep{lu2024rtllm} score the
digital code a model writes, and AnalogCoder \citep{lai2025analogcoder} the analog
circuits it designs.  Others test circuit understanding: CIRCUIT
\citep{skelic2025circuit} grades the numerical answers to 510 analog-circuit
questions, AMSbench \citep{shi2025amsbench} asks about 8,000 multimodal questions on
schematic perception, analysis, and design, and MMCircuitEval
\citep{zhao2025mmcircuiteval} 3,614 on digital and analog circuits across the design
flow, while CircuChain \citep{ravishankara2026circuchain} pairs circuit-analysis
problems with variants that invert the usual sign conventions to test whether models
follow the stated conventions or their training priors.  Razavi-Bench complements them in three ways: its 50 questions were written by an
analog designer, Razavi, who also graded AI answers to them, and we calibrate our
reference answers and grader against his verdicts; it follows 73 models released over
30 months on the same questions; and its look-alike circuits separate models that
analyze a drawn circuit from those that recognize its shape.
Razavi-Bench turns the 50 questions into a benchmark and applies it to 73 models
(Figure~\ref{fig:timeline}).  Four findings stand out:

\begin{enumerate}[leftmargin=*,itemsep=2pt]
\item \textbf{The top score rose from 28.83\% to 98.33\% in 30 months,} mostly
before the reference answers were published (Section~\ref{sec:progress}).
\item \textbf{Textbook circuits are nearly solved; look-alikes are not.}  The
hardest questions show a circuit that resembles a familiar one but works
differently, or ask about behavior the obvious analysis misses
(Section~\ref{sec:hard}).
\item \textbf{Models identify a circuit before they analyze it.}  The strongest
models then check the individual devices; most others keep the familiar name
(Section~\ref{sec:recognize}).
\item \textbf{The last points are the expensive ones,} but the cost of reaching
90\% fell 17-fold within five months (Section~\ref{sec:cost}).
\end{enumerate}

\section{Benchmark and Evaluation Setup}
\label{sec:setup}

Each model receives one message holding the question text and, where needed, the
schematic, cropped to the part the question asks about and without any title that
would name the circuit.  There is no system prompt and no tool or web access.
Models run at the highest reasoning effort each offers and at its maximum output
length, and answer every question in three independent trials.  The Overall score
is the share of available rubric points awarded.  Cost covers all 150 answers at
OpenRouter's list prices on November 10, 2026, excluding grading.

GPT-6 Luna, at maximum reasoning effort, grades each answer against the reference
answer and rubric \citep{zheng2023judging}, from 0 points (incorrect) to 4 (correct
conclusion with correct reasoning).  Its grades agree closely with Razavi's own on
the AI answers he graded in the two articles: its totals match his for Part 1 and
fall within three points for Part 2.  The agreement is partly by construction,
since we refined the references using his verdicts; with those removed, its totals
fall a few points below his.  All questions, rubrics, answers, and grades are
public on the project page.

\section{Progress Over Time}
\label{sec:progress}

The first questions were published in November 2025, and the full reference
answers on June 30, 2026.  About 40\% of the improvement came before the first of
these dates and nearly 90\% before the second, so most of the gain cannot have come
from training on the reference answers.

Ten models now score at least 90\%.  The three leaders answer nearly every question
fully correctly and never score zero.  The top two, Gemini 4 Carbon and Claude Opus
5.5, are within trial-to-trial variation of each other (their trials span 98.0\% to
98.5\% and 97.0\% to 98.5\%), and both are clearly ahead of the rest: the worst
trial of each beats every other model's best.

Release date correlates with score, but recent models still vary widely, largely
with price: among models released since July 2026, the median score rises from
about 65\% for those costing under \$1 for the 150 answers to about 90\% for those
costing more than \$10.

\section{Which Questions Are Hard, and Why?}
\label{sec:hard}

\subsection{The easiest and hardest questions}

Table~\ref{tab:questions} lists the easiest and hardest questions.  The easiest ask
about single-device facts or textbook concepts, which even the oldest models mostly
answer correctly.  Circuits that must be recognized from the drawing were hard until
recently: no 2024 model recognized a plain source follower, whereas nearly every
recent model analyzes it fully.

\begin{table}[t]
\centering
\caption{The easiest and hardest questions, by mean score over the 219 answers to
each (73 models $\times$ 3 trials).  P1 and P2 denote Part 1 and Part 2.}
\label{tab:questions}
\small
\begin{tabularx}{\textwidth}{@{}l X r r@{}}
\toprule
Key & Question & Mean score & Full credit \\
\midrule
\multicolumn{4}{@{}l}{\emph{Easiest}} \\
P1 Q2 & Two students disagree on whether a MOSFET's transconductance rises or falls with overdrive voltage; who is right?  (Both, depending on what is held constant) & 99.0\% & 96\% \\
P1 Q14 & Analyze the CMOS inverter of Figure~\ref{fig:lookalike}(c) (open-ended) & 97.5\% & 95\% \\
P1 Q3 & Is a PMOS device's small-signal model identical to an NMOS device's? & 97.3\% & 93\% \\
P1 Q1 & What happens to a MOSFET's intrinsic gain if its length and width double? & 93.7\% & 85\% \\
P1 Q22 & Explain why the Miller effect is less pronounced in a cascode & 92.2\% & 72\% \\
\midrule
\multicolumn{4}{@{}l}{\emph{Hardest}} \\
P1 Q21 & Is the circuit of Figure~\ref{fig:lookalike}(b) a cascode?  (No: a PMOS source follower drives a common-gate stage) & 23.1\% & 21\% \\
P1 Q15 & Analyze the two-NMOS circuit of Figure~\ref{fig:lookalike}(d) (open-ended) & 26.3\% & 24\% \\
P2 Q14 & Role of an inverter added across the master latch of a divide-by-two circuit (a feedforward path, not a keeper) & 31.6\% & 9\% \\
P2 Q15 & Is that added inverter a keeper, as a quoted answer claims?  (No) & 33.2\% & 7\% \\
P2 Q5 & Small-signal resistance looking into a ring oscillator's supply node & 33.7\% & 15\% \\
P2 Q19 & How a capacitor added at an LC oscillator's tail node changes phase noise & 37.7\% & 2\% \\
P2 Q6 & Does the supply-node resistance of P2 Q5 change if the ring oscillates?  (Yes) & 38.8\% & 5\% \\
P1 Q6 & Can a PMOS device with gate and drain grounded act as a current source?  (No: it is diode-connected) & 39.4\% & 27\% \\
\bottomrule
\end{tabularx}
\end{table}

The hardest questions fall into two groups.  The first shows a circuit or device
that resembles a familiar one but works differently: the CMOS inverter and the
cascode redrawn with a changed transistor type (Section~\ref{sec:recognize}), a
diode-connected transistor offered as a current source, and an inverter in a
divide-by-two circuit that looks like a latch keeper but actually forms a
feedforward path.  Models tend to answer the familiar circuit rather than the drawn
one; even Claude Fable 5 reads the keeper in every trial.

The second group asks how a circuit behaves where the obvious analysis does not
apply.  Part 2 questions 5 and 6 ask for the small-signal resistance looking into
the supply node of a ring oscillator.  Treating each inverter as two
diode-connected devices gives $(1/3)(1/g_{mn}+1/g_{mp})$, but the ring oscillates,
and the resistance is then set by the charging of the node capacitances, about
$1/(3f_0C_L)$.  Most answers give only the static result, and even when the next
question states that result and asks whether oscillation changes it, few answers
get it fully right.  A capacitor added at an LC oscillator's tail node (Part 2
question 19) can raise phase noise in one range of values, lower it in another, and
cause low-frequency instability when it is too large.  Most answers give only one
direction, usually that the capacitor helps.

The leader's few remaining errors are mostly tradeoffs, such as widening a
comparator's cross-coupled pair, which speeds the latch up only until the added
capacitance takes over.  It makes no errors on the look-alikes.

Part 2 is harder on average than Part 1, yet models that answer without extended
reasoning score higher on Part 2.  Part 2 asks how performance changes, which a
model can partly recall from textbooks, whereas Part 1 mostly asks it to work out
how a drawn circuit operates, and that is where reasoning pays off.

\subsection{Recognizing or analyzing?}
\label{sec:recognize}

\begin{figure}[t]
\centering
\includegraphics[width=\textwidth]{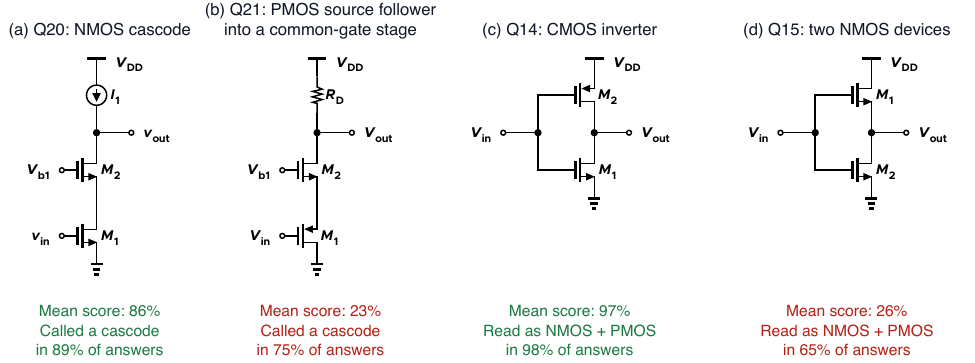}
\caption{Two textbook circuits (green) and their look-alikes (red), from Part 1
questions 20, 21, 14, and 15, with each question's mean score and, as read by GPT-6
Luna, how the answers identify the circuit.}
\label{fig:lookalike}
\end{figure}

Does a model analyze a circuit, or recognize its shape and recall the familiar one?
Two pairs of Part 1 questions tell these apart (Figure~\ref{fig:lookalike}).
Question 21 shows the cascode of question 20, except that the lower transistor is a
PMOS device with its drain grounded, so it acts as a source follower driving the
common-gate transistor.  Question 15 shows the CMOS inverter of question 14 with two
NMOS devices, so the upper one is a source follower.

Answers that call a look-alike by its textbook name always score zero, and those
that see through it average more than three of the four points.  Newer models fall
for the look-alikes less often, but many still do; the top models almost never do.

Reasoning traces show the order of the steps.  Most first name a familiar circuit,
often before stating any transistor type, even when the question does not name it.
For textbook circuits this shortcut works; on the look-alikes the outcome depends on
whether the model later tests the name against the symbols.  The failures take two
forms.  On the cascode look-alike, about half the failing answers never discuss the
transistor types; on the inverter look-alike, most failing traces work out the
transistor types from the symbols and still keep the familiar name, and even those
that later drop it recover only partial credit.  Two traces show the difference:

\begin{quote}\small
\emph{Claude Fable 5.2 (4 points):} ``Now I'm checking M1's arrow at its top
terminal, which marks it as a PMOS with its source at the top, tied to M2's source,
its drain grounded, and its gate at Vin.  That makes M1 a source follower feeding
M2's source, so the stack is not a cascode.''

\emph{Claude Opus 4.8 (0 points):} ``Checking the transistor symbols, the arrows
don't settle the NMOS question, but the topology is clear regardless.\ \ldots{}
That's the textbook cascode configuration, so the answer is yes.''
\end{quote}

The strongest models check consistently and get both look-alikes right in every
trial.  Below them, seeing through one look-alike does not mean seeing through the
other.  Because many traces are provider summaries, which may compress the order of
the steps, this ordering is indicative rather than exact; the full, unsummarized
traces, mostly from mid-ranked models, show it even more starkly.

\section{What Does Cost Buy?}
\label{sec:cost}

\begin{figure}[t]
\centering
\includegraphics{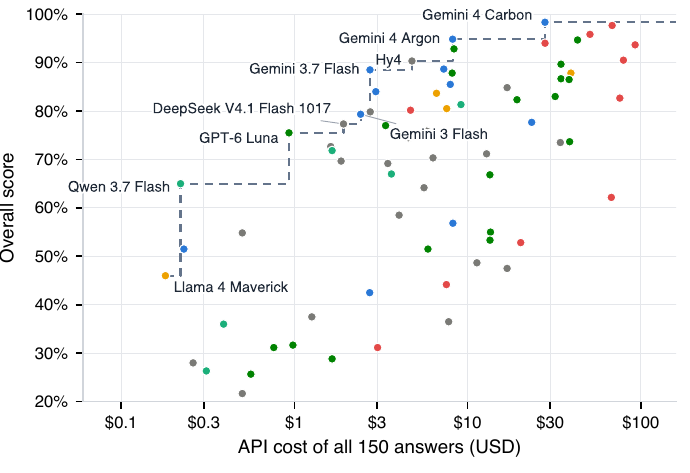}
\caption{Overall score against the API cost of all 150 answers at OpenRouter's list
prices (logarithmic axis).  The dashed line is the Pareto front; colors are as in
Figure~\ref{fig:timeline}.}
\label{fig:cost}
\end{figure}

Figure~\ref{fig:cost} plots score against cost.  Along its Pareto front, each step
up costs more than the last: about 88\% costs under \$3 for the 150 answers, about
95\% about \$8, and the last 3.5 points to the top score another \$20.

Gemini 4 Carbon pays for its lead in tokens, writing about 3.5 times as many per
answer as Gemini 4 Argon at the same price per token.  The most expensive models,
recent Claude Fable and Opus versions, cost more than twice as much as Gemini 4
Carbon without beating it.  Cost and score correlate, even among recent models.

\begin{table}[t]
\centering
\caption{The cheapest model reaching each Overall score, in the month the score was
first reached and in November 2026.  Costs are for all 150 answers.}
\label{tab:cheapest}
\small
\begin{tabularx}{\textwidth}{@{}l X X r@{}}
\toprule
Score & Cheapest when first reached & Cheapest in November 2026 & Reduction \\
\midrule
$\geq$60\% & Gemini 3 Flash, Dec 2025, \$2.41 & Qwen 3.7 Flash, Jul 2026, \$0.22 & 11$\times$ \\
$\geq$70\% & Gemini 3 Flash, Dec 2025, \$2.41 & GPT-6 Luna, Sep 2026, \$0.93 & 2.6$\times$ \\
$\geq$80\% & GPT-5.3, Feb 2026, \$19.27 & Gemini 3.7 Flash, Aug 2026, \$2.73 & 7.1$\times$ \\
$\geq$90\% & Claude Fable 5, Jun 2026, \$79.33 & Hy4, Nov 2026, \$4.77 & 17$\times$ \\
\bottomrule
\end{tabularx}
\end{table}

A correct answer has become only somewhat cheaper.  On the questions answered with
full credit both before 2026 and since July 2026, the median full-credit answer
costs \$0.039 from the recent models against \$0.052 from the older ones, because
cheaper models now answer correctly; within each
question, the cost of a full-credit answer has risen, as answers grew longer faster
than tokens became cheaper.  What has fallen sharply is the cost of reaching a given
score (Table~\ref{tab:cheapest}).  Nearly two-thirds of the questions can now be
answered reliably for about a cent each; money matters only on the rest.

\section{Conclusion}
\label{sec:conclusion}

The errors that remain are of two kinds.  Many models answer from a circuit's
familiar shape, and outside the strongest models, checking the name against the
transistor symbols is inconsistent: a model that sees through one look-alike can
fall for another.  Separately, the leader's remaining losses are on tradeoffs that
have no one-directional answer.  Benchmarks that aim to measure circuit analysis
rather than recall should therefore include both look-alike circuits and such
tradeoffs on purpose.

Razavi-Bench also has clear limits: its 50 questions are few, short, and
self-contained, its leader is within two points of the maximum, and its grades come
from a single language-model judge.  Analog design means turning a specification
into a circuit, simulating it, and revising it; tracking progress from here will
need agentic design benchmarks such as Analog Design
Bench~\citep{analogdesignbench2026} and, beyond them, tasks that are harder, more
varied, and closer to industrial design practice.

\FloatBarrier
\bibliographystyle{iclr2027_conference}
\bibliography{references}

\end{document}